\documentclass[Journal,SingleSpace,NoLineNumbers,InsideFigs,NoLists,letterpaper]{ascelike-new}
\usepackage{amsmath,amssymb}
\usepackage{graphicx}
\usepackage{booktabs}
\usepackage{array}
\usepackage{float}
\usepackage[section]{placeins}
\usepackage{enumitem}
\usepackage[authoryear,round]{natbib}
\bibpunct{(}{)}{;}{a}{~}{,}
\usepackage[hidelinks]{hyperref}
\newcommand{\doilink}[1]{doi: \url{https://doi.org/#1}}

\setlist[itemize]{leftmargin=1.8em,itemsep=0.15em,topsep=0.2em}
\setlist[enumerate]{leftmargin=1.8em,itemsep=0.15em,topsep=0.2em}

\usepackage{microtype}

\title {Joint-State Inference for Autonomous Excavators Using Cab-View Vision--Control Fusion}
\author[1]{Junwei Li}
\author[2,6]{Jian Pan\textsuperscript{*}}
\author[3]{Mengchi Cai}
\author[4]{Shuai Zhao}
\author[2,6]{Xizhuo Jiang}
\author[2]{Zheng Wang}
\author[5]{Qing Xu}
\author[5]{Keqiang Li}
\author[5]{Jianqiang Wang}
\affil[1]{National School of Elite Engineers, Northeastern University, Shenyang, Liaoning, China.}
\affil[2]{School of Mechanical Engineering and Automation, Northeastern University, Shenyang, Liaoning, China.}
\affil[3]{School of Civil Engineering, Tsinghua University, Beijing, China.}
\affil[4]{College of Information, Liaoning University, Shenyang, Liaoning, China.}
\affil[5]{School of Vehicle and Mobility, Tsinghua University, Beijing, China.}
\affil[6]{State Key Laboratory of Intelligent Deep Metal Mining and Equipment, Shenyang 110819, China.}

\begin{document}
\maketitle
\begin{center}
\footnotesize\textsuperscript{*}Corresponding author. Email: panjian@me.neu.edu.cn
\end{center}

\begin{abstract}
Reliable state feedback of excavator front-end mechanisms is essential for autonomous excavation, yet retrofitting existing machines with dedicated joint sensors, external cameras, or visual markers can increase deployment and maintenance burdens. This paper proposes a closed-loop autonomous excavation framework for continuous inference of excavator joint states using onboard visual observations and control-related temporal information. The framework encodes structural cues from the current cab-view image, models short-term motion trends from historical joystick commands, and introduces a rolling state buffer for the bucket joint to preserve temporal continuity under severe occlusion. Mechanical joint-range constraints are incorporated into the learning objective to suppress infeasible estimates. The current cab-view image and inferred states are supplied to an Action Chunking with Transformers (ACT)-based control policy for online action generation. Experiments in an AGX Dynamics for Unity simulation environment under diverse operating conditions evaluate joint-state inference and end-to-end task execution. Across 20 autonomous excavation trials, the framework achieved a 95\% task-completion rate, with an average joint-state-inference-and-control latency of 10.11~ms per step. The bucket-joint inference achieved a mean absolute error (MAE) of $0.965^{\circ}$. These results demonstrate a sensor-light and computationally efficient joint-state-inference approach that can support closed-loop autonomous excavation with reduced dependence on additional sensing infrastructure.

\end{abstract}

\section{Introduction}

Excavation operations are commonly conducted in unstructured and highly disturbed environments. Although traditional manual operation can accomplish complex tasks by relying on the experience of skilled operators, long-term continuous excavation requires substantial human involvement and relies heavily on operator expertise. Autonomous excavation technology provides a promising approach for improving excavation safety, operational efficiency, and standardization \citep{zhang2021autonomous}. However, achieving continuous and stable autonomous operation places stringent requirements on joint-state inference accuracy, decision-making efficiency, and execution reliability. Therefore, a reliable joint-state inference and autonomous control framework with reduced dependence on additional state-sensing devices is a fundamental prerequisite for high-quality autonomous excavation \citep{dadhich2016challenges}.

Existing autonomous excavation systems typically rely on additional state-sensing devices to measure excavator joint states, which provide critical feedback for motion planning, process adjustment, and task execution \citep{lee2022teleoperation,kim2020task}. Such devices, including hydraulic cylinder displacement sensors, inertial measurement units (IMUs), and inclination sensors, can directly measure excavator states. However, their deployment generally requires hardware installation and system modification, increasing deployment cost and structural-integration complexity, as well as requirements for calibration, maintenance, and fault diagnosis. Moreover, excavators often operate under dust, vibration, and mechanical impacts, which may compromise the measurement stability and long-term reliability of these devices. Therefore, continuous acquisition of excavator joint states without extensive hardware modification, followed by their use as state feedback for autonomous operation, is important for the intelligent upgrading of existing excavators \citep{ito2021effects}.

Previous research on excavator state acquisition has mainly focused on marker-assisted measurement and vision-based perception for estimating the working-mechanism pose. Marker-assisted methods can provide high-precision pose measurements under controlled conditions. However, they depend strongly on specific equipment configurations and experimental setups, requiring marker installation and manual configuration on excavators. These constraints hinder rapid deployment across excavator platforms, while the marker systems increase configuration complexity and maintenance cost in practical applications \citep{soltani2017skeleton}. Vision-based joint-state estimation methods typically use externally mounted fixed cameras to capture excavation images and infer working-mechanism states through object detection, keypoint extraction, or visual reconstruction. Although these approaches reduce mechanical modification, they are constrained by camera placement, viewing geometry, and field conditions, which limits their use in changing work areas and long-term autonomous operation \citep{guo2021computer,roberts2020vision,nakamura2020pose}. Moreover, severe occlusion can require complex visual-processing pipelines or multi-stage optimization, increasing computational cost and hindering continuous state feedback for real-time closed-loop control. Existing approaches therefore still struggle to combine accurate state acquisition, convenient deployment, and real-time closed-loop operation.

\begin{figure}[tbp]
    \centering
    \begin{minipage}[b]{0.24\textwidth}
        \centering
        \includegraphics[width=\textwidth,height=0.8\textwidth]{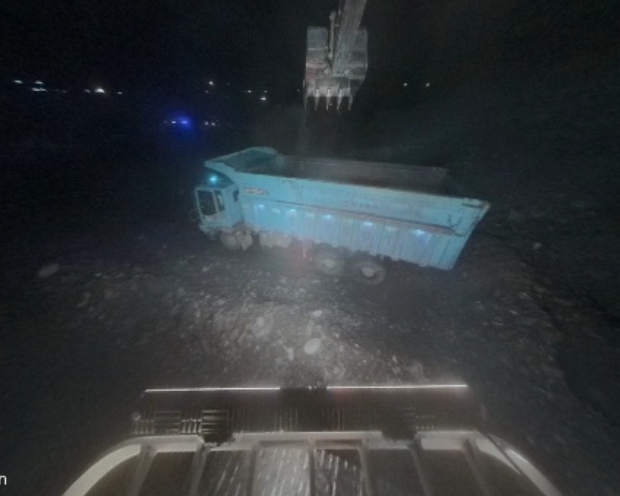}\\[-0.25em]
        \small\textbf{(a)}
    \end{minipage}\hfill
    \begin{minipage}[b]{0.24\textwidth}
        \centering
        \includegraphics[width=\textwidth,height=0.8\textwidth]{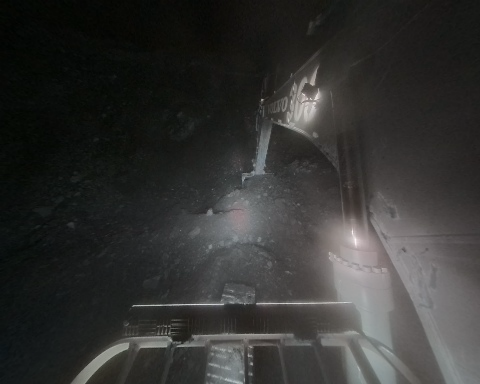}\\[-0.25em]
        \small\textbf{(b)}
    \end{minipage}\hfill
    \begin{minipage}[b]{0.24\textwidth}
        \centering
        \includegraphics[width=\textwidth,height=0.8\textwidth]{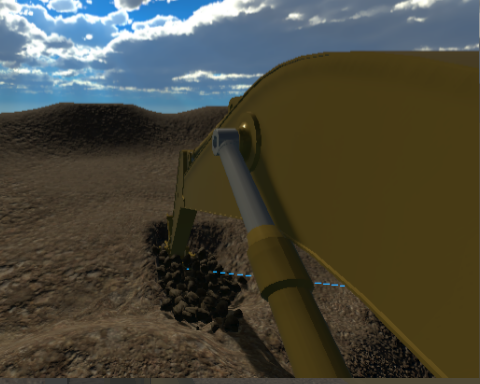}\\[-0.25em]
        \small\textbf{(c)}
    \end{minipage}\hfill
    \begin{minipage}[b]{0.24\textwidth}
        \centering
        \includegraphics[width=\textwidth,height=0.8\textwidth]{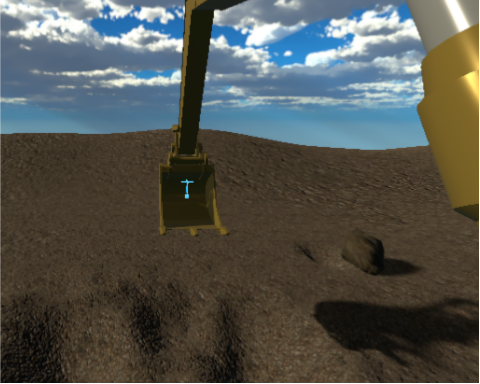}\\[-0.25em]
        \small\textbf{(d)}
    \end{minipage}
    \caption{Representative cab-view scenes in real and simulated excavation settings. (a)--(b) are collected from the publicly available Fuxi excavator-motion dataset \citep{fuxi_excavator_motion}, which provides synchronized RGB observations and joint-state measurements, whereas (c)--(d) are generated from the in-house AGX Dynamics for Unity simulation dataset. The bucket and distal working-mechanism structures can be partially or fully obscured during digging and loading, reducing the reliability of single-frame joint-state inference.}
    \label{fig:occlusion-cases}
\end{figure}

Representative cab-view scenes from real and simulated excavation settings, including typical partial and severe working-mechanism occlusions, are shown in Fig.~\ref{fig:occlusion-cases}.

To address the limitations of existing autonomous excavation systems, including their dependence on additional state-sensing devices, high modification costs, and incomplete closed-loop operation, this paper proposes a low-cost, robust closed-loop autonomous excavation framework. The framework replaces the conventional device-dependent state-measurement paradigm with a lightweight state inference module that continuously estimates excavator joint states. It establishes a complete workflow from joint-state inference and real-time feedback to autonomous decision-making and action execution. The main contributions of this paper are summarized as follows:
\begin{enumerate}
    \item A perception--execution-coupled closed-loop autonomous excavation framework is developed, integrating joint-state inference, real-time feedback, autonomous decision-making, and task execution into a unified pipeline. The framework enables continuous and stable excavation under the tested simulated excavation conditions, supporting the intelligent upgrading of excavators.
    \item A state inference module that does not require additional state-sensing devices is designed. It uses intrinsically available excavator information to estimate excavator joint states, reducing dependence on additional sensing hardware and the costs of intelligent retrofitting and long-term maintenance.
    \item Extensive simulation experiments validate the framework. The results demonstrate low joint-state inference and control-execution latency, a high task-completion rate, and a high demonstration-normalized excavation yield. The inferred joint states are also robust to visual appearance variations, including illumination and component-color changes.
\end{enumerate}

\section{Background and Related Work}

\subsection{Joint-State Acquisition in Autonomous Excavation}
Autonomous excavation has attracted increasing attention in the field of intelligent construction because it can improve operational efficiency, safety, and productivity in complex construction environments. Unlike conventional industrial robots operating in structured scenarios, excavators perform long-horizon interaction tasks in highly unstructured environments, involving complex body motion, continuous contact with surrounding materials, and dynamic interaction with the workspace. Therefore, accurate and continuous inference of excavator joint states is fundamental to autonomous perception, decision-making, and closed-loop control \citep{parascho2023construction,nguyen2023robotic,bock2015future}.

Existing excavator state acquisition approaches mainly rely on onboard contact-based sensing systems, including joint encoders, hydraulic cylinder displacement sensors, and inertial measurement units \citep{eraliev2022sensing}. For example, \citet{jud2021heap} developed an autonomous walking excavator system that integrated joint states, inertial measurements, and environmental perception information to achieve state estimation and autonomous operation control. \citet{tang2023fullbody} further fused multiple onboard sensing modalities to estimate the full-body pose of excavators. These approaches establish direct measurement relationships between mechanical components and their corresponding states, providing reliable feedback for motion analysis and control. However, deploying additional sensing systems on existing excavators usually requires mechanical modification, sensor calibration, and long-term maintenance, which increases system complexity and limits their large-scale application on commercial construction equipment.

Recent advances in deep learning-based visual perception, including convolutional feature extraction, keypoint detection, and pose-regression methods, have demonstrated the feasibility of recovering spatial states from monocular observations. Representative approaches, such as residual networks, visual localization networks, and structure-aware keypoint-estimation frameworks, have significantly improved image-based robotic perception \citep{he2016deep,kendall2015posenet,newell2016hourglass,pavlakos2017coarse}. Motivated by these advances, low-cost vision-based joint-state inference methods have gradually been introduced into construction machinery.

\subsection{Marker-assisted and Markerless Pose Estimation}
However, directly recovering excavator joint states from visual observations remains challenging because of complex mechanical motion and severe self-occlusion. The excavator working mechanism consists of multiple articulated components that create complex occlusion relationships across operating stages. In particular, the bucket is frequently invisible during digging and curling, creating substantial uncertainty between visual observations and the true joint states \citep{xu2019vision}.

To alleviate the limitations of visual observations, some studies introduced additional visual constraints or external observations to assist excavator pose estimation. Marker-based approaches establish explicit correspondences between image features and mechanical components by attaching artificial markers to the working mechanism. \citet{lundeen2016optical} developed an optical marker-based pose estimation system for articulated excavators, where visual markers were deployed to recover the three-dimensional motion of the excavator end-effector. \citet{feng2018camera} further proposed a camera-marker network for articulated machine pose estimation, integrating multiple cameras and visual markers to improve the measurement accuracy of complex mechanical systems.

Although these approaches can improve pose-estimation accuracy, they introduce deployment requirements such as marker installation, external-camera calibration, and auxiliary observation systems. These requirements increase system complexity and reduce the feasibility of large-scale deployment on existing excavators. Moreover, these approaches mainly focus on accurate pose measurement under controlled conditions, whereas their ability to provide continuous state feedback for autonomous control in natural operating environments remains insufficiently explored.

To reduce dependence on artificial assistance, markerless vision-based estimation methods have been increasingly explored. \citet{liang2019vision} proposed a markerless pose estimation framework for articulated construction robots, which directly detected robot keypoints from images through deep neural networks and further recovered two-dimensional and three-dimensional poses. This study demonstrated the potential of learning-based visual perception for construction machinery, while also revealing that severely occluded components, such as buckets, remain difficult to estimate reliably. Subsequently, \citet{mahmood2022synthetic} trained convolutional neural networks using synthetic images to improve data availability for excavator three-dimensional pose estimation. \citet{assadzadeh2022vision} further constructed hybrid datasets combining real and synthetic data to enhance model generalization under different scenarios.

Despite these advances, existing vision-based excavator joint-state estimation methods mainly focus on offline pose-reconstruction accuracy, with performance commonly evaluated using single-frame images or pre-collected datasets. The temporal stability of inferred joint states and their usability as state feedback for autonomous control have received comparatively limited attention. In practical excavation tasks, the working mechanism evolves continuously, and visual occlusion or observation loss can interrupt joint-state inference. Therefore, current visual observations alone may be insufficient to maintain stable state feedback \citep{liu2024vision}. Exploiting historical observations and temporal correlations to improve joint-state inference continuity and control usability under partial observability therefore remains an important challenge for autonomous excavation systems.

\section{Computational Framework}

\subsection{Problem Definition}
In autonomous excavation scenarios, continuous and reliable joint-state feedback is essential for closed-loop decision-making and autonomous execution. Existing autonomous excavation approaches typically rely on additional state-sensing devices to obtain excavator joint states. However, these devices increase the hardware deployment cost and system-integration complexity of intelligent retrofitting, while reducing the adaptability of autonomous systems across excavator platforms. Therefore, continuous inference of key excavator joint states without additional state-sensing devices, followed by their use as state feedback for autonomous operation, remains a critical challenge for low-cost, reliable autonomous excavation.

To address this challenge, this work reformulates excavator joint-state acquisition as an online joint-state inference problem based on intrinsically available excavator information. Specifically, the proposed approach uses cab-view images, joystick inputs, and historical state information collected during excavation operations to continuously infer the current joint states and provide them as feedback for autonomous control. Let $I_t^c$ denote the cab-view image at time $t$, and let $\mathbf{u}_t$ denote the joystick input. The excavator joint-state vector is defined as
\[
\mathbf{q}_t = [
q_t^{\mathrm{swing}},
q_t^{\mathrm{boom}},
q_t^{\mathrm{stick}},
q_t^{\mathrm{bucket}}].
\]
where the four elements represent the swing, boom, stick, and bucket joint states, respectively. The objective is to infer the current joint state $\hat{\mathbf{q}}_t$ without relying on additional state-sensing devices and provide the inferred state as online feedback for an autonomous excavation controller.

Considering the heterogeneous motion characteristics and visibility conditions of different excavator joints, joint-specific temporal representations are constructed. For the swing, boom, and stick joints, which generally maintain better visual observability during operation, the current cab-view image and historical joystick inputs are used as the primary information sources:
\begin{equation}
\left(I_t^c,\mathbf{U}_{t-1}\right)\rightarrow
\left[
\hat{q}_t^{\mathrm{swing}},
\hat{q}_t^{\mathrm{boom}},
\hat{q}_t^{\mathrm{stick}}
\right],
\end{equation}
where $\mathbf{U}_{t-1}=\{\mathbf{u}_{t-T},\ldots,\mathbf{u}_{t-1}\}$ denotes the historical joystick-input sequence over the preceding $T$ time steps. In contrast, the bucket joint frequently suffers from severe visual ambiguity during digging and loading. Therefore, an additional historical state-feedback mechanism is introduced:
\begin{equation}
\left(I_t^c,\hat{\mathbf{Q}}_{t-1},\mathbf{U}_{t-1}\right)
\rightarrow \hat{q}_t^{\mathrm{bucket}},
\end{equation}
where $\hat{\mathbf{Q}}_{t-1}=\{\hat{\mathbf{q}}_{t-T},\ldots,\hat{\mathbf{q}}_{t-1}\}$ represents the historical inferred joint states maintained during online inference. During offline evaluation and closed-loop execution, the first $T=10$ frames of each trajectory are inferred from cab-view images to initialize the state-history buffer. Thereafter, the buffer is updated recursively using the model's inferred joint states, and no ground-truth joint states are used as inputs.

\subsection{Overview of the Autonomous Excavation Framework}

This paper proposes a closed-loop autonomous excavation framework for continuous excavator joint-state inference and autonomous task execution without additional state-sensing devices. The framework consists of a state inference module and an autonomous execution module. By establishing a closed-loop information flow of "joint-state inference--state feedback--action generation--execution update", the framework realizes a complete pipeline from joint-state inference to autonomous excavation task completion.

\begin{figure}[htbp]
    \centering
    \includegraphics[width=\textwidth,trim=0 44bp 0 0,clip]{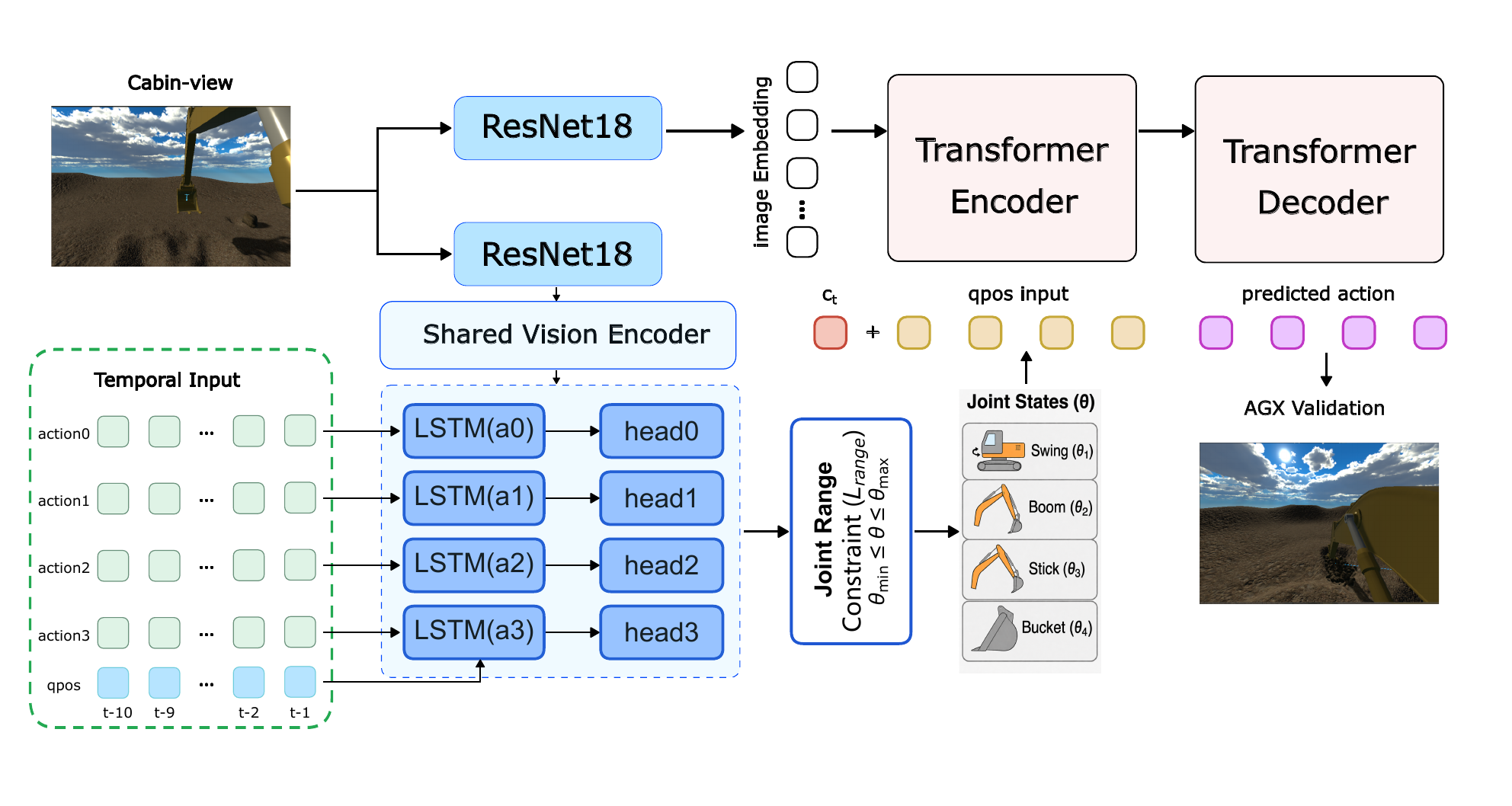}
    \caption{Overview of the proposed closed-loop autonomous excavation framework. Cab-view images, joystick-input history, and a rolling state buffer support joint-state inference; the image, inferred states, and task context are provided to ACT for action generation.}
    \label{fig:framework}
\end{figure}

Fig.~\ref{fig:framework} summarizes the information flow and coupling between the state-inference and autonomous-execution modules.

The inferred joint states generated by the state inference module provide state feedback to the autonomous controller and are combined with the current cab-view image and task objectives for action generation. In this work, an Action Chunking with Transformers (ACT)-based control policy \citep{zhao2023icmlw-learning} serves as the autonomous execution module. It generates continuous control actions from the current multimodal observations and drives the excavator to complete target excavation tasks. The inferred state at the current time step is added to the historical state buffer to provide temporal information for subsequent joint-state inference. Through this recursive process, the framework establishes a closed-loop connection among joint-state inference, feedback update, and action execution. Compared with conventional approaches that rely on external state-measurement devices, the framework reduces dependence on additional state-sensing hardware, improves deployment flexibility on existing excavator platforms, and provides a feasible solution for low-cost intelligent upgrading.

\subsection{Mechanism-Constrained Excavator State Inference Module}
The state inference module establishes a unified representation by exploiting complementary information from visual observations, joystick inputs, and historical motion evolution.

The current cab-view image is encoded by a visual backbone $\phi_c(\cdot)$:
\begin{equation}
\mathbf{f}_t^c=\phi_c(I_t^c).
\end{equation}
where $\mathbf{f}_t^c$ denotes the extracted visual representation. It captures the spatial configuration and appearance characteristics of visible excavator components, including the boom, stick, bucket, and machine body.
To reduce online computational overhead, only the current image frame is processed. Temporal information is implicitly introduced through historical control signals and state feedback rather than requiring long image sequences.

Since different joints are driven by different control channels and exhibit distinct motion patterns, independent temporal encoders are designed for each joint. The historical joystick inputs are encoded and integrated with the visual representation:
\begin{equation}
\begin{aligned}
\mathbf{h}_t^{\mathrm{swing}} &= \mathcal{T}_{\mathrm{swing}}(\mathbf{f}_t^c,\mathbf{U}_{t-1}^{\mathrm{swing}}),\\
\mathbf{h}_t^{\mathrm{boom}} &= \mathcal{T}_{\mathrm{boom}}(\mathbf{f}_t^c,\mathbf{U}_{t-1}^{\mathrm{boom}}),\\
\mathbf{h}_t^{\mathrm{stick}} &= \mathcal{T}_{\mathrm{stick}}(\mathbf{f}_t^c,\mathbf{U}_{t-1}^{\mathrm{stick}}).
\end{aligned}
\end{equation}
For the bucket joint, a historical state feedback branch is introduced:
\begin{equation}
\begin{aligned}
\mathbf{h}_t^{\mathrm{bucket}} &= \mathcal{T}_{\mathrm{bucket}}(\mathbf{f}_t^c,\hat{\mathbf{Q}}_{t-1},\mathbf{U}_{t-1}^{\mathrm{bucket}}).
\end{aligned}
\end{equation}
where $\mathcal{T}_i(\cdot)$ denotes the temporal modeling module specific to joint $i$, and $\mathbf{U}_{t-1}^{i}$ denotes the preceding joystick-input history associated with the control channel of joint $i$.
Finally, independent regression heads are employed to infer each joint state:
\begin{equation}
\hat{q}_t^i=g_i(\mathbf{h}_t^i).
\end{equation}
where $g_i(\cdot)$ denotes the regression head for joint $i$, and $i\in\{\mathrm{swing},\mathrm{boom},\mathrm{stick},\mathrm{bucket}\}$.

To encourage temporally consistent predictions, each training trajectory segment is initialized with a state-history buffer comprising the ground-truth joint states from the preceding $T=10$ frames. The model is then rolled out autoregressively for $K=10$ steps. At each rollout step, the predicted joint state is appended to the state-history buffer, and the oldest state is discarded. The updated buffer is used to infer the subsequent joint state. The regression and joint-range losses are accumulated over all rollout steps. The joint regression loss is defined as:
\begin{equation}
L_q=\frac{1}{K}\sum_{j=0}^{K-1}\frac{1}{n}\sum_{i=1}^{n}
\left(\hat{q}_{t+j}^{\,i}-q_{t+j}^{\,i}\right)^2,
\end{equation}
where $K=10$ represents the autoregressive rollout length, $n=4$ represents the number of inferred joints, and $q_i^{\min}$ and $q_i^{\max}$ denote the lower and upper physical operating limits of joint $i$, respectively. Although learning-based models can approximate joint states from data, unconstrained predictions may occasionally violate mechanical limits. Therefore, a joint-range regularization term is introduced:
\begin{equation}
L_{\mathrm{range}}=\frac{1}{K}\sum_{j=0}^{K-1}\sum_{i=1}^{n}
\left[
\max(0,\hat{q}_{t+j}^{\,i}-q_i^{\max})+
\max(0,q_i^{\min}-\hat{q}_{t+j}^{\,i})
\right].
\end{equation}
The final optimization objective is:
\begin{equation}
L=\lambda_q L_q+\lambda_r L_{\mathrm{range}}.
\end{equation}
where $\lambda_q$ and $\lambda_r$ control the relative contributions of the joint-regression and joint-range regularization terms, respectively. The proposed constraint embeds intrinsic mechanical priors into the learning objective through joint operating limits, enhancing the physical consistency of inferred joint states and improving the reliability of state feedback for autonomous control.

\subsection{Closed-Loop Autonomous Excavation Execution}

To connect joint-state inference with autonomous task execution, the framework integrates the state inference module with an ACT-based control policy to form a state-feedback-driven online closed-loop process. Rather than requiring direct access to ground-truth joint states, the policy takes the current cab-view image, inferred joint states, and task context as inputs and generates continuous action sequences. Unlike conventional control strategies that rely on manually designed rules or complex trajectory planning, ACT learns the state--action mapping from expert demonstrations and directly produces executable control actions. Benefiting from its action-chunking structure, ACT avoids additional online planning and provides efficient inference, making it suitable for real-time autonomous excavation tasks.
\begin{equation}
\mathbf{A}_t=\pi_{\mathrm{ACT}}(I_t^c,\hat{\mathbf{q}}_t,\mathbf{c}_t),
\qquad
\mathbf{A}_t=\{\mathbf{a}_t,\ldots,\mathbf{a}_{t+H-1}\},
\label{eq:act-policy}
\end{equation}
where $\mathbf{c}_t$ denotes the task context supplied to the policy and $H=100$ is the action-chunk horizon. Following the standard ACT temporal-ensembling protocol, the policy is queried at every control cycle, and predictions for the current control instant from overlapping action chunks are combined by $\mathrm{TE}(\cdot)$ to obtain the executable action $\tilde{\mathbf{a}}_t$. Only $\tilde{\mathbf{a}}_t$, rather than the entire action chunk $\mathbf{A}_t$, is transmitted to the excavator simulator; the next observation is then processed and the state-feedback--action-generation procedure is repeated. Thus, ACT serves as a policy interface that converts the latest cab-view image and inferred joint state into temporally coherent control actions, while the state inference module continuously refreshes the state feedback on which those actions depend. For the online protocol, $\mathcal{O}_t=\{I_t^c,\mathbf{U}_{t-1}\}$ denotes the exogenous observation available at time $t$, whereas $\hat{\mathbf{Q}}_{t-1}$ is the internal rolling buffer of preceding inferred joint states. The complete state-inference module is denoted by $f_\theta(\cdot)$, and $\mathcal{E}(\cdot)$ denotes the simulator state-transition function induced by the executed action. The resulting online execution procedure is summarized in Table~\ref{tab:online-protocol}.

\begin{table}[tbp]
\centering
\caption{Online closed-loop execution procedure.}
\label{tab:online-protocol}
\small
\setlength{\tabcolsep}{0pt}
\begin{tabular}{@{}l@{\hspace{1.7em}}l@{\hspace{0.35em}}c@{\hspace{0.35em}}l@{}}
\toprule
Process & \multicolumn{3}{c}{Online operation} \\
\midrule

Observation &
 $\mathcal{O}_t$ & $\leftarrow$ & $\{I_t^c,\mathbf{U}_{t-1}\}$
\\[6pt]

State inference &
 $\hat{\mathbf{q}}_t$ & $\leftarrow$ & $f_{\theta}(I_t^c,\mathbf{U}_{t-1},\hat{\mathbf{Q}}_{t-1})$
\\[6pt]

Action generation &
 $\mathbf{A}_t$ & $\leftarrow$ & $\pi_{\mathrm{ACT}}(I_t^c,\hat{\mathbf{q}}_t,\mathbf{c}_t)$
\\[6pt]

Temporal ensembling &
 $\tilde{\mathbf{a}}_t$ & $\leftarrow$ & $\mathrm{TE}(\mathbf{A}_{t-H+1},\ldots,\mathbf{A}_{t})$
\\[6pt]

Excavator execution &
 $\mathbf{q}_{t+1}$ & $\leftarrow$ & $\mathcal{E}(\tilde{\mathbf{a}}_t)$
\\[6pt]

State feedback &
 $\hat{\mathbf{Q}}_{t}$ & $\leftarrow$ & $\mathrm{Append}(\hat{\mathbf{Q}}_{t-1},\hat{\mathbf{q}}_t)$
\\

\bottomrule
\end{tabular}
\end{table}

For ACT policy training, 50 expert demonstration trajectories, each containing 2,000 control steps, were used. The trajectories were split into 40 training trajectories and 10 validation trajectories. At each control cycle, the ACT policy received multimodal observations comprising the current cab-view image $I_t^c$, the four-dimensional inferred joint-state vector $\hat{\mathbf{q}}_t$, and task context. The task context contained seven task-stage features, including a six-dimensional one-hot stage encoding and a scalar stage-progress value, together with a three-dimensional target excavation position. Thus, the numerical state-and-context input was 14-dimensional, whereas the cab-view image was processed as a separate visual input and fused with the numerical features for action generation. Each control action was four-dimensional, $\mathbf{a}_t=[a_t^{\mathrm{swing}},a_t^{\mathrm{boom}},a_t^{\mathrm{stick}},a_t^{\mathrm{bucket}}]^{\mathsf{T}}$.

The state inference module comprised a ResNet-18 visual encoder pretrained on ImageNet and joint-specific LSTM temporal encoders. Cab-view images and joystick inputs were acquired at 30 Hz. All cab-view images were resized to $224 \times 224$ pixels before being input to the network, and the preceding $T=10$ joystick-control inputs were used as temporal input. Each LSTM encoder comprised two layers with a hidden dimension of 128. For the bucket branch, which is frequently affected by visual occlusion during excavation, a rolling buffer of inferred joint states was additionally maintained to provide temporal context and stabilize state inference under partial observability. The state-inference model was trained by minimizing the joint-state regression loss and the physical-range constraint loss, with $\lambda_q=1$ and $\lambda_r=0.2$. The allowable joint ranges were set according to the physical motion limits of the excavator model. Adam optimization was used with a batch size of 8 for up to 200 epochs, and the model with the lowest validation MAE was selected for evaluation. Unless otherwise specified, the same training configuration was used for all state-inference variants. The ACT-based controller operated at the same 30-Hz control frequency and received the current cab-view image, inferred joint-state vector, and task context at each control cycle.

\section{Experimental Design}

\subsection{State-Inference Evaluation Protocol}

The experiments evaluate the proposed autonomous excavation framework along four dimensions: continuous joint-state inference, the contributions of different information sources, robustness to visual appearance variations, and support for closed-loop autonomous excavation.

A visual-only joint-state inference experiment is conducted on a public excavator dataset, using a model that receives only cab-view images as a comparison method. This experiment investigates whether visual observations alone can support excavator joint-state inference and analyzes the limitations of visual information under occlusion and visual ambiguity. The results motivate the subsequent inclusion of joystick inputs and historical state feedback.

The complete state inference module is evaluated on an in-house AGX Dynamics for Unity simulation dataset. An autonomous excavation scenario is constructed in an AGX Dynamics for Unity simulation environment for data collection and framework validation. The dataset is collected through human teleoperation and contains synchronized cab-view images, excavator joint states, and joystick inputs. A total of 50 excavation episodes are recorded, with 2,000 simulation steps per episode. The sampling frequency is 30 Hz, and each time step contains an aligned image--state--control tuple. The 50 episodes were split by complete trajectories into 40 training episodes, 5 validation episodes, and 5 test episodes. The validation episodes were used for checkpoint selection based on MAE. The test episodes were held out from state-inference model training and checkpoint selection and were used exclusively for offline evaluation of state-inference performance. This dataset is used to analyze the contributions of visual information, historical joystick inputs, and rolling state feedback to continuous joint-state inference. By progressively incorporating these information sources, the influence of each component on joint-state inference accuracy and stability is quantified.

To evaluate robustness under visual appearance variations, appearance-perturbation experiments are conducted. The excavator mechanical structure, motion process, and control-input distribution are kept unchanged, while illumination and working-component colors are modified. By separating appearance changes from mechanical and behavioral variations, these experiments analyze the dependence of the state inference module on specific texture features and color distributions and evaluate its adaptability under diverse visual conditions. For joint $i$ with ground-truth angle $q_t^i$ and inferred angle $\hat{q}_t^i$ at time step $t$, joint-state inference performance is evaluated using mean absolute error (MAE), root mean square error (RMSE), and maximum absolute error $D_{\max}$:
\begin{equation}
\mathrm{MAE}_i=\frac{1}{N}\sum_{t=1}^{N}|\hat{q}_t^i-q_t^i|,
\quad
\mathrm{RMSE}_i=\sqrt{\frac{1}{N}\sum_{t=1}^{N}(\hat{q}_t^i-q_t^i)^2},
\quad
D_{\max,i}=\max_{1\leq t\leq N}|\hat{q}_t^i-q_t^i|.
\end{equation}
Here, $N$ is the number of evaluated time steps. MAE and RMSE quantify average inference error, whereas $D_{\max,i}$ captures the largest transient error for joint $i$.
A complete closed-loop autonomous excavation evaluation is conducted in the AGX Dynamics for Unity simulation environment. Unlike offline experiments that focus on numerical joint-state inference accuracy, this evaluation verifies whether inferred joint states can support autonomous control through continuous, reliable state feedback. During the experiment, the excavator completes a full excavation sequence, including approaching, cutting, curling, lifting, swinging, and dumping. Overall framework performance is evaluated using task-completion rate, demonstration-normalized excavation yield, and computational latency.

\subsection{Closed-Loop Evaluation Protocol}

The same ACT-based control policy was evaluated using ground-truth and inferred joint states. For each pair of trials, the two state-input conditions were independently simulated from identical initial conditions, task objectives, soil distributions, and random seeds; only the joint-state input source differed. Each state-input condition was evaluated over 20 paired trials to assess the effect of the state-input source on autonomous excavation performance.

A trial was considered successful when the excavator completed the predefined excavation procedure and the excavated soil mass exceeded the predefined threshold of $m_{\mathrm{thr}}=600\,\mathrm{kg}$. This threshold was selected to represent an effective soil excavation process rather than marginal contact or limited soil disturbance. The task-completion rate was calculated as
\begin{equation}
R_{\mathrm{success}}=
\frac{N_{\mathrm{success}}}{N_{\mathrm{total}}}
\times 100\%,
\end{equation}
where $N_{\mathrm{success}}$ denotes the number of successful trials and $N_{\mathrm{total}}$ denotes the total number of trials.

We define the demonstration-normalized excavation yield as the ratio of the total excavated soil mass under state-input mode $s$ to the total excavated soil mass from the demonstration executions. Let $M_{\mathrm{demo}}$ denote the total excavated soil mass obtained from 20 demonstration executions, and let $M_s$ denote the total excavated soil mass obtained from 20 closed-loop executions under state-input mode $s$. It was calculated as
\begin{equation}
\eta_s=
\frac{M_s}{M_{\mathrm{demo}}}
\times 100\%,
\end{equation}
where $s\in\{\mathrm{ground\ truth},\mathrm{inferred}\}$.

\section{Results}

\subsection{Analysis of Visual Observation for State Inference}

The Fuxi dataset is first used to investigate the capability and limitations of cab-view visual observations for excavator joint-state inference. The results show that image-based perception provides sufficient visual cues for joints with relatively stable visibility, enabling the visual-only inference model to capture the overall motion trends of the boom, stick, and bucket.

However, because of partial occlusion and limited visual observability during excavation operations, the bucket joint exhibits increased prediction uncertainty. Without additional motion-related information, the visual-only inference model suffers from local fluctuations, temporal inconsistency, and degraded joint-state inference accuracy when visual evidence becomes insufficient. These observations demonstrate that cab-view perception provides valuable state information but requires complementary temporal and control-related priors for reliable state feedback in autonomous excavation scenarios.

\begin{figure}[tbp]
    \centering
    \begin{minipage}[t]{0.39\textwidth}
        \centering
        \textbf{(a)}\par\vspace{0.15em}
        \includegraphics[width=\textwidth]{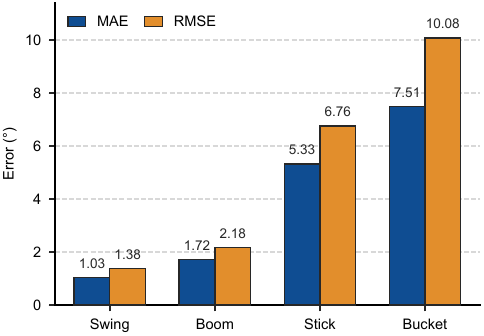}
    \end{minipage}\hfill
    \begin{minipage}[t]{0.56\textwidth}
        \centering
        \textbf{(b)}\par\vspace{0.15em}
        \includegraphics[width=\textwidth]{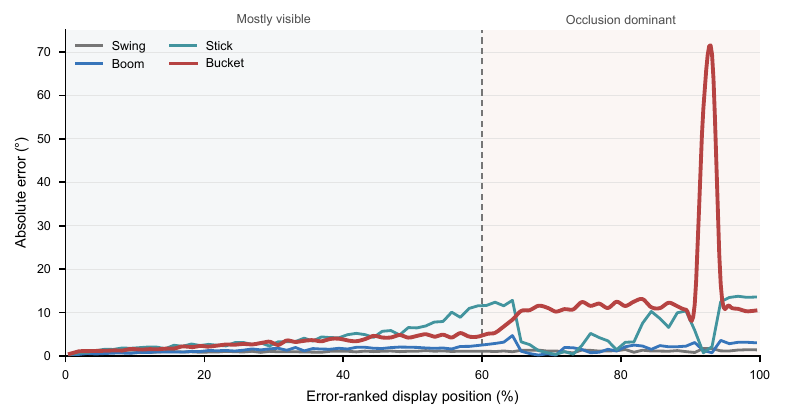}
    \end{minipage}
    \setlength{\Abovecaptionrule}{0.40in}
    \caption{Visual-only joint-state inference errors on the public Fuxi dataset. (a) Joint-wise MAE and RMSE. (b) Absolute error profiles ordered by display position, with the mostly visible and occlusion-dominant ranges indicated; the horizontal axis denotes ranked display position rather than time.}
    \label{fig:fuxi}
\end{figure}

Fig.~\ref{fig:fuxi} quantifies these visibility-dependent error patterns for visual-only joint-state inference.

\subsection{Evaluation of State Inference Performance on the Simulation Dataset}

The proposed framework is evaluated on an in-house AGX Dynamics for Unity simulation dataset containing diverse excavator trajectories. As summarized in Table~\ref{tab:joint-metrics-sim}, it achieves accurate and stable joint-state inference for all four joints, with predictions closely matching the ground-truth trajectories.

The swing, boom, and stick joints achieve low inference errors, indicating that visual structural cues and joint-specific control dynamics sufficiently constrain these observable components. Although the bucket joint remains challenging because of frequent visual ambiguity, historical state feedback improves temporal consistency and reduces abrupt fluctuations, providing more reliable feedback for downstream autonomous control.
\begin{table}[tbp]
    \centering
    \caption{Joint-level prediction metrics on the simulation dataset}
    \label{tab:joint-metrics-sim}
    \small
    \renewcommand{\arraystretch}{1.12}
    \begin{tabular*}{0.72\textwidth}{@{\extracolsep{\fill}}lccc@{}}
        \toprule
        Joint & MAE ($^\circ$) & RMSE ($^\circ$) & $D_{\max}$ ($^\circ$) \\
        \midrule
        Swing  & 0.694 & 0.868 & 2.787 \\
        Boom   & 0.018 & 0.048 & 0.793 \\
        Stick  & 0.128 & 0.213 & 2.382 \\
        Bucket & 0.965 & 1.550 & 7.083 \\
        \bottomrule
    \end{tabular*}
\end{table}

Table~\ref{tab:ablation} compares several input configurations. Historical states alone do not consistently improve performance because recursive errors may accumulate. Joystick inputs reduce the maximum deviations of the swing, boom, and stick joints but increase aggregate MAE ($9.470^\circ$ versus $4.374^\circ$ for visual-only inference) and remain insufficient for the visually ambiguous bucket joint. Combining historical state feedback and control priors reduces the maximum bucket deviation from $23.424^\circ$ to $7.083^\circ$. The joint-range constraint prevents predictions from exceeding physical operating limits rather than directly improving accuracy.

\begin{figure}[tbp]
    \centering
    \includegraphics[width=0.88\textwidth]{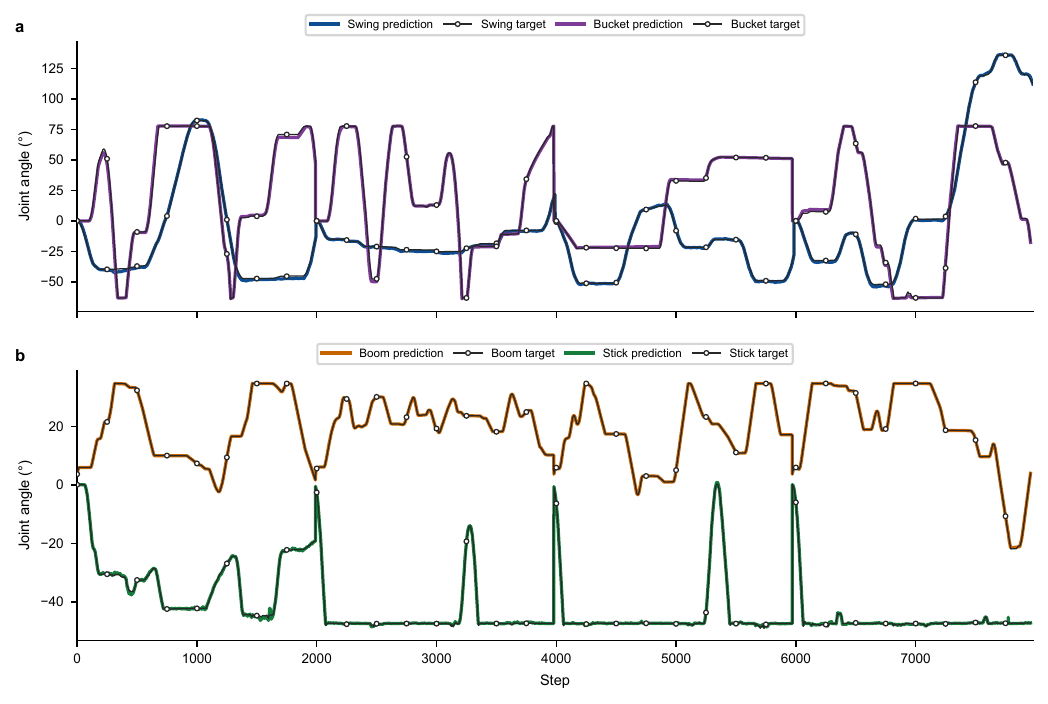}
    \setlength{\Abovecaptionrule}{0.40in}
    \caption{Target and predicted trajectories for the four joints over a representative 8,000-step segment of the simulation test data. The upper panel shows swing and bucket, whereas the lower panel shows boom and stick. Colored curves denote predictions; dark-gray curves with open-circle markers denote target angles.}
    \label{fig:joint-overview}
\end{figure}

Fig.~\ref{fig:joint-overview} compares the target and predicted joint trajectories obtained with the proposed configuration.

\begin{table}[tbp]
    \centering
    \caption{Comparison of input configurations (all values in degrees)}
    \label{tab:ablation}
    \small
    \renewcommand{\arraystretch}{1.12}
    \begin{tabular*}{\textwidth}{@{\extracolsep{\fill}}p{0.29\textwidth}ccccc@{}}
        \toprule
        Method & MAE & Bucket $D_{\max}$ & Swing $D_{\max}$ & Boom $D_{\max}$ & Stick $D_{\max}$ \\
        \midrule
        ResNet & 4.374 & 23.424 & 7.193 & 4.042 & 8.817 \\
        ResNet + historical states & 31.632 & 35.483 & 20.530 & 26.294 & 30.983 \\
        ResNet + control inputs & 9.470 & 21.560 & \textbf{2.787} & \textbf{0.793} & \textbf{2.382} \\
        ResNet + states + controls & \textbf{0.451} & \textbf{7.083} & - & - & - \\
        \bottomrule
    \end{tabular*}
\end{table}

\subsection{Robustness to Visual Appearance Variations}

\begin{figure}[tbp]
    \centering
    \includegraphics[width=0.24\textwidth]{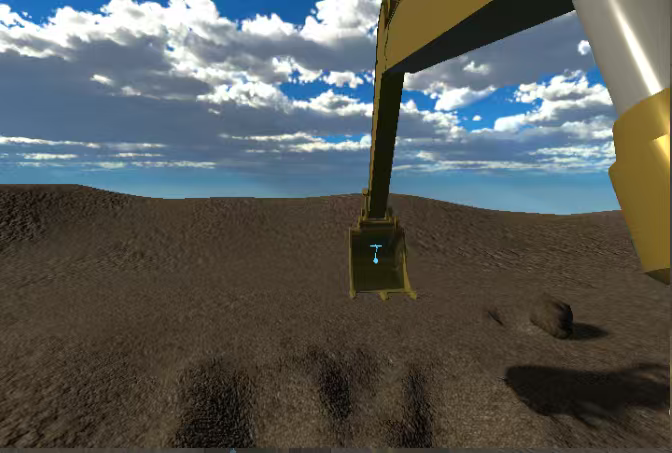}\hfill
    \includegraphics[width=0.24\textwidth]{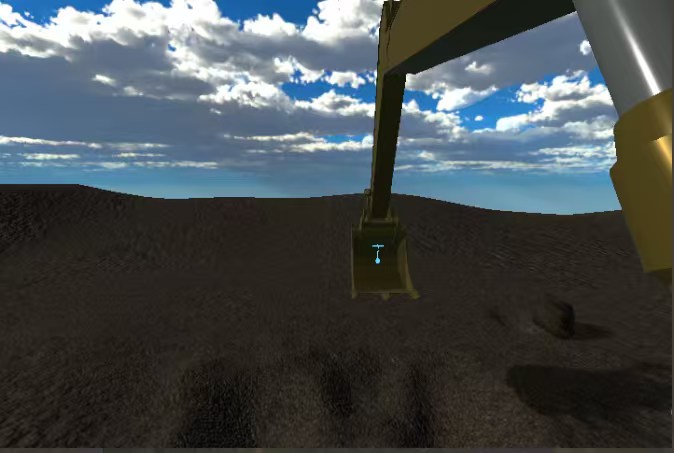}\hfill
    \includegraphics[width=0.24\textwidth]{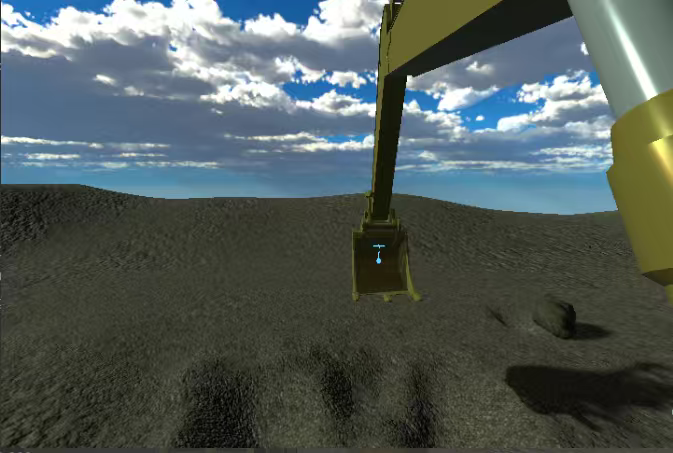}\hfill
    \includegraphics[width=0.24\textwidth]{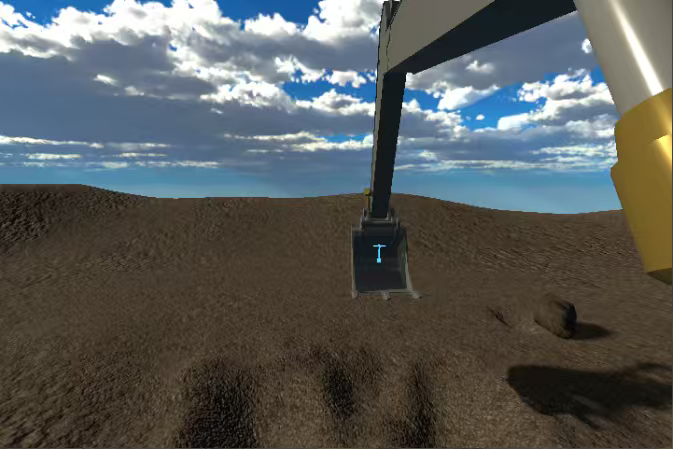}
    \caption{Appearance perturbation settings: initial condition, two lighting changes, and a representative color-change condition.}
    \label{fig:generalization-scenes}
\end{figure}

To evaluate the robustness of the proposed state inference module against visual appearance variations, lighting conditions and component colors are modified while keeping the working-mechanism configuration, operation sequence, and control-input distribution unchanged. Because the bucket joint suffers from relatively severe visual occlusion during excavation, it is selected as a representative case for appearance-robustness analysis. Fig.~\ref{fig:generalization-scenes} illustrates the perturbation settings, and Table~\ref{tab:generalization} summarizes the corresponding bucket-joint inference errors.

Under different lighting conditions, the bucket-joint MAE increases from $0.965^\circ$ in the original setting to $1.072^\circ$ and $1.013^\circ$, respectively, corresponding to relative increases of $11.09\%$ and $4.97\%$. Meanwhile, component-color variations result in MAEs of $0.946^\circ$ and $1.055^\circ$, remaining comparable with the original configuration. These results demonstrate that the proposed state inference module maintains stable joint-state inference performance under moderate visual appearance variations, suggesting that its predictions are not solely dependent on specific textures or color distributions.

\begin{table}[tbp]
    \centering
    \caption{Bucket prediction error under appearance perturbations}
    \label{tab:generalization}
    \small
    \renewcommand{\arraystretch}{1.12}
    \begin{tabular*}{0.86\textwidth}{@{\extracolsep{\fill}}lccc@{}}
        \toprule
        Condition & MAE ($^\circ$) & RMSE ($^\circ$) & Relative MAE Change (\%) \\
        \midrule
        Initial condition & 0.965 & 1.550 & Baseline \\
        Lighting change 1 & 1.072 & 1.704 & +11.09\% \\
        Lighting change 2 & 1.013 & 1.683 & +4.97\% \\
        Component color 1 & 0.946 & 1.519 & -1.97\% \\
        Component color 2 & 1.055 & 1.691 & +9.33\% \\
        \bottomrule
    \end{tabular*}
\end{table}

\subsection{Closed-Loop Autonomous Excavation Performance}

When inferred joint states are provided as feedback to the ACT-based control policy \citep{zhao2023icmlw-learning}, the simulated excavator executes the complete excavation sequence, including approaching, cutting, curling, lifting, swinging, and dumping. The inferred-state condition completed 19 of 20 trials (95\%) and achieved a demonstration-normalized excavation yield of 93\%, whereas the ground-truth-state condition completed all 20 trials and achieved a demonstration-normalized excavation yield of 98\%. These results demonstrate that the proposed state inference module provides temporally consistent state feedback suitable for downstream autonomous excavation control, rather than only improving offline joint-state prediction accuracy.

Table~\ref{tab:closed-loop-comparison} compares task-level performance and per-step latency when the same ACT-based control policy receives ground-truth or inferred joint states. Latency was measured on a workstation equipped with an AMD Ryzen 9 9950X3D processor and an NVIDIA GeForce RTX 5070 Ti GPU. Inference-based state feedback requires 10.11 ms per control step, compared with 8.24 ms for direct ground-truth state input, corresponding to an additional latency of 1.87 ms. The state inference module itself takes approximately 1.87 ms, indicating that the framework maintains low computational overhead within the AGX Dynamics for Unity closed-loop control pipeline.

\begin{table}[tbp]
    \centering
    \caption{Closed-loop task performance using ground-truth and inferred joint-state inputs.}
    \label{tab:closed-loop-comparison}
    \small
    \setlength{\tabcolsep}{2pt}
    \begin{tabular*}{0.88\textwidth}{@{}>{\raggedright\arraybackslash}p{0.23\textwidth}@{\extracolsep{\fill}}>{\centering\arraybackslash}p{0.11\textwidth}>{\centering\arraybackslash}p{0.14\textwidth}>{\centering\arraybackslash}p{0.16\textwidth}>{\centering\arraybackslash}p{0.10\textwidth}@{}}
        \toprule
        State input to ACT & Successful trials & Completion rate (\%) & Demonstration-normalized excavation yield (\%) & Latency (ms) \\
        \midrule
        Ground-truth joint states & 20/20 & 100 & 98 & 8.24 \\
        Inferred joint states & 19/20 & 95 & 93 & 10.11 \\
        \bottomrule
    \end{tabular*}
\end{table}

\FloatBarrier

\section{Discussion}

The primary contribution of this work lies in establishing a closed-loop autonomous excavation framework under limited sensing conditions, in which excavator joint-state inference is formulated as a structured, rolling, controller-oriented process. By exploiting available onboard information, the framework reduces dependence on additional state-sensing devices and associated calibration procedures, providing a sensor-light solution for autonomous excavator deployment. Historical joystick inputs provide motion priors during partial observability, while joint-range constraints improve the physical consistency of inferred joint states before they are delivered to downstream controllers. Offline joint-state inference evaluation, robustness analysis, and closed-loop excavation experiments collectively demonstrate the feasibility of the framework for low-cost autonomous excavation.

Despite these encouraging results, several limitations remain. First, although the proposed framework is validated in an AGX Dynamics for Unity simulation environment, real-excavator experiments under construction-site conditions have not yet been conducted. Transfer from simulation to physical machines may introduce additional challenges, including sensor noise, mechanical uncertainty, and environmental disturbances. Second, the appearance-perturbation experiments mainly consider illumination and component-color variations, whereas more complex factors such as soil contamination, camera degradation, vibration-induced image blur, weather conditions, and severe occlusion require further investigation. Third, the closed-loop evaluation focuses on task-level performance, including task-completion rate, demonstration-normalized excavation yield, and computational latency. Future work will investigate large-scale real-world evaluation, failure-mode analysis, and more adaptive joint-state inference strategies to further improve the reliability and generalization of autonomous excavation systems.
\section{Conclusions}

This paper proposes a closed-loop autonomous excavation framework aimed at enabling continuous excavator joint-state inference and autonomous task execution. Unlike methods that rely solely on isolated visual observations, the proposed framework integrates cab-view images with control-related temporal information to provide continuous joint-state feedback under partial visual observability. By coupling a lightweight state-inference module with an ACT-based control policy, the framework establishes a complete closed-loop process from onboard state perception to online autonomous action generation.
\section*{Data Availability Statement}
Some or all data, models, or code that support the findings of this study are available from the corresponding author upon reasonable request.

\section*{Acknowledgments}
This work was supported by the Applied Basic Research Program of Liaoning Province (Grant No.~2025JH2/101330110), the National Natural Science Foundation of China (Grant No.~52404162), and the Independent Research Project of the State Key Laboratory of Intelligent Deep Metal Mining and Equipment (No.~IDMEIRI2508).

\section*{Conflicts of Interest}
The authors declare no known competing financial interests or personal relationships that could have appeared to influence the work reported in this paper.


\begin{thebibliography}{Zhao et~al. (2023)}
\small
\bibitem[Assadzadeh et~al.(2023)]{assadzadeh2022vision} Assadzadeh, A., M. Arashpour, H. Li, R. Hosseini, F. Elghaish, and S. Baduge. 2023. ``Excavator 3D pose estimation using deep learning and hybrid datasets.'' \textit{Advanced Engineering Informatics}, 55:101875. \doilink{10.1016/j.aei.2023.101875}.
\bibitem[Bock(2015)]{bock2015future} Bock, T. 2015. ``The future of construction automation: Technological disruption and the upcoming ubiquity of robotics.'' \textit{Automation in Construction}, 59:113--121. \doilink{10.1016/j.autcon.2015.07.022}.
\bibitem[Dadhich et~al.(2016)]{dadhich2016challenges} Dadhich, S., U. Bodin, and U. Andersson. 2016. ``Key challenges in automation of earth-moving machines.'' \textit{Automation in Construction}, 68:212--222. \doilink{10.1016/j.autcon.2016.05.009}.
\bibitem[Eraliev et~al.(2022)]{eraliev2022sensing} Eraliev, O. M. U., K. H. Lee, D. Y. Shin, and C. H. Lee. 2022. ``Sensing, perception, decision, planning and action of autonomous excavators.'' \textit{Automation in Construction}, 141:104428. \doilink{10.1016/j.autcon.2022.104428}.
\bibitem[Feng et~al.(2018)]{feng2018camera} Feng, C., V. R. Kamat, and H. Cai. 2018. ``Camera marker networks for articulated machine pose estimation.'' \textit{Automation in Construction}, 96:148--160. \doilink{10.1016/j.autcon.2018.09.004}.
\bibitem[Fuxi Robot(2025)]{fuxi_excavator_motion} Fuxi Robot. 2025. \textit{Excavator-motion} [Dataset]. Hugging Face. \url{https://huggingface.co/datasets/fuxi-robot/excavator-motion} (accessed August 7, 2026).
\bibitem[Guo et~al.(2021)]{guo2021computer} Guo, B. H. W., Y. Zou, Y. H. Fang, Y. M. Goh, and P. X. W. Zou. 2021. ``Computer vision technologies for safety science and management in construction: A critical review and future research directions.'' \textit{Safety Science}, 135:105130. \doilink{10.1016/j.ssci.2020.105130}.
\bibitem[He et~al.(2016)]{he2016deep} He, K., X. Zhang, S. Ren, and J. Sun. 2016. ``Deep residual learning for image recognition.'' \textit{Proc., IEEE Conf. on Computer Vision and Pattern Recognition}, 770--778. \doilink{10.1109/CVPR.2016.90}.
\bibitem[Ito et~al.(2021)]{ito2021effects} Ito, M., C. Raima, S. Saiki, Y. Yamazaki, and Y. Kurita. 2021. ``Effects of machine instability feedback on safety during digging operation in teleoperated excavators.'' \textit{IEEE Access}, 9:28987--28998. \doilink{10.1109/ACCESS.2021.3059710}.
\bibitem[Jud et~al.(2021)]{jud2021heap} Jud, D., S. Kerscher, M. Wermelinger, E. Jelavic, P. Egli, P. Leemann, G. Hottiger, and M. Hutter. 2021. ``HEAP---The autonomous walking excavator.'' \textit{Automation in Construction}, 129:103783. \doilink{10.1016/j.autcon.2021.103783}.
\bibitem[Kendall et~al.(2015)]{kendall2015posenet} Kendall, A., M. Grimes, and R. Cipolla. 2015. ``PoseNet: A convolutional network for real-time 6-DOF camera relocalization.'' \textit{Proc., IEEE Int. Conf. on Computer Vision}, 2938--2946. \doilink{10.1109/ICCV.2015.336}.
\bibitem[Kim et~al.(2020)]{kim2020task} Kim, J., D. Lee, and J. Seo. 2020. ``Task planning strategy and path similarity analysis for an autonomous excavator.'' \textit{Automation in Construction}, 112:103108. \doilink{10.1016/j.autcon.2020.103108}.
\bibitem[Lee et~al.(2022)]{lee2022teleoperation} Lee, J. S., Y. Ham, H. Park, and J. Kim. 2022. ``Challenges, tasks, and opportunities in teleoperation of excavator toward human-in-the-loop construction automation.'' \textit{Automation in Construction}, 135:104119. \doilink{10.1016/j.autcon.2021.104119}.
\bibitem[Liang et~al.(2019)]{liang2019vision} Liang, C. J., K. M. Lundeen, W. McGee, C. C. Menassa, S. Lee, and V. R. Kamat. 2019. ``A vision-based marker-less pose estimation system for articulated construction robots.'' \textit{Automation in Construction}, 104:80--94. \doilink{10.1016/j.autcon.2019.04.004}.
\bibitem[Liu et~al.(2024)]{liu2024vision} Liu, G., Q. Wang, T. Wang, B. Li, and X. Xi. 2024. ``Vision-based excavator pose estimation for automatic control.'' \textit{Automation in Construction}, 157:105162. \doilink{10.1016/j.autcon.2023.105162}.
\bibitem[Lundeen et~al.(2016)]{lundeen2016optical} Lundeen, K. M., S. Dong, N. Fredricks, M. Akula, J. Seo, and V. R. Kamat. 2016. ``Optical marker-based end effector pose estimation for articulated excavators.'' \textit{Automation in Construction}, 65:51--64. \doilink{10.1016/j.autcon.2016.02.003}.
\bibitem[Mahmood et~al.(2022)]{mahmood2022synthetic} Mahmood, B., S. Han, and J. Seo. 2022. ``Implementation experiments on convolutional neural network training using synthetic images for 3D pose estimation of an excavator on real images.'' \textit{Automation in Construction}, 133:103996. \doilink{10.1016/j.autcon.2021.103996}.
\bibitem[Nakamura et~al.(2020)]{nakamura2020pose} Nakamura, H., Y. Tsukada, T. Tamaki, B. Raytchev, and K. Kaneda. 2020. ``Pose estimation of excavators.'' \textit{Proc., Int. Workshop on Advanced Imaging Technology}, 115152J. \doilink{10.1117/12.2566436}.
\bibitem[Newell et~al.(2016)]{newell2016hourglass} Newell, A., K. Yang, and J. Deng. 2016. ``Stacked hourglass networks for human pose estimation.'' \textit{Proc., European Conf. on Computer Vision}, 483--499. \doilink{10.1007/978-3-319-46484-8_29}.
\bibitem[Nguyen and Ha(2023)]{nguyen2023robotic} Nguyen, H. A. D., and Q. P. Ha. 2023. ``Robotic autonomous systems for earthmoving equipment operating in volatile conditions and teaming capacity: A survey.'' \textit{Robotica}, 41(2):486--510. \doilink{10.1017/S0263574722000339}.
\bibitem[Parascho(2023)]{parascho2023construction} Parascho, S. 2023. ``Construction robotics: From automation to collaboration.'' \textit{Annual Review of Control, Robotics, and Autonomous Systems}, 6:183--204. \doilink{10.1146/annurev-control-080122-090049}.
\bibitem[Pavlakos et~al.(2017)]{pavlakos2017coarse} Pavlakos, G., X. Zhou, K. G. Derpanis, and K. Daniilidis. 2017. ``Coarse-to-fine volumetric prediction for single-image 3D human pose.'' \textit{Proc., IEEE Conf. on Computer Vision and Pattern Recognition}, 7025--7034. \doilink{10.1109/CVPR.2017.139}.
\bibitem[Roberts et~al.(2020)]{roberts2020vision} Roberts, D., W. T. Calderon, S. Tang, and M. Golparvar-Fard. 2020. ``Vision-based construction worker activity analysis informed by body posture.'' \textit{Journal of Computing in Civil Engineering}, 34:04020017. \doilink{10.1061/(ASCE)CP.1943-5487.0000898}.
\bibitem[Soltani et~al.(2017)]{soltani2017skeleton} Soltani, M. M., Z. H. Zhu, and A. Hammad. 2017. ``Skeleton estimation of excavator by detecting its parts.'' \textit{Automation in Construction}, 82:1--15. \doilink{10.1016/j.autcon.2017.06.023}.
\bibitem[Tang et~al.(2023)]{tang2023fullbody} Tang, J., M. Wang, H. Luo, P. K. Y. Wong, X. Zhang, W. Chen, and J. C. P. Cheng. 2023. ``Full-body pose estimation for excavators based on data fusion of multiple onboard sensors.'' \textit{Automation in Construction}, 147:104694. \doilink{10.1016/j.autcon.2022.104694}.
\bibitem[Xu and Yoon(2019)]{xu2019vision} Xu, J., and H. S. Yoon. 2019. ``Vision-based estimation of excavator manipulator pose for automated grading control.'' \textit{Automation in Construction}, 98:122--131. \doilink{10.1016/j.autcon.2018.11.022}.
\bibitem[Zhang et~al.(2021)]{zhang2021autonomous} Zhang, L., J. Zhao, P. Long, et al. 2021. ``An autonomous excavator system for material loading tasks.'' \textit{Science Robotics}, 6(55):eabc3164. \doilink{10.1126/scirobotics.abc3164}.
\bibitem[Zhao et~al.(2023)]{zhao2023icmlw-learning} Zhao, T. Z., V. Kumar, S. Levine, and C. Finn. 2023. ``Learning fine-grained bimanual manipulation with low-cost hardware.'' \textit{Proc., Robotics: Science and Systems XIX}. \doilink{10.15607/rss.2023.xix.016}.
\end{thebibliography}
\end{document}